\documentclass{article}

\usepackage{microtype}
\usepackage{graphicx}
\usepackage{subfigure}
\usepackage{booktabs} 
\usepackage{amsmath}
\usepackage{amssymb}
\usepackage{amsfonts}
\usepackage{array}
\usepackage[ruled,vlined,linesnumbered,algo2e]{algorithm2e}
\usepackage{xcolor}
\usepackage{lipsum}
\usepackage{comment}
\usepackage{siunitx}
\usepackage{graphicx}
\usepackage{float}
\usepackage{appendix}
\usepackage{hyperref}
\usepackage{optidef}
\usepackage{adjustbox}
\usepackage{nameref}
\usepackage{fancyhdr}
\usepackage{titlesec}
\usepackage{hhline}
\usepackage{array, boldline, rotating}
\usepackage{afterpage}

\usepackage{hyperref}

\newcommand{\cut}[1]{}

\usepackage{multirow}

\usepackage[accepted]{icml2019}

\icmltitlerunning{Measuring the Transferability of Adversarial Examples}

\begin{document}
\twocolumn[
\icmltitle{Measuring the Transferability of Adversarial Examples}




\begin{icmlauthorlist}
\icmlauthor{Deyan V. Petrov}{ed}
\icmlauthor{Timothy M. Hospedales}{ed}
\end{icmlauthorlist}

\icmlaffiliation{ed}{School of Informatics, University of Edinburgh, Edinburgh, United Kingdom}

\icmlcorrespondingauthor{Deyan Petrov}{deqn1996@gmail.com}
\icmlcorrespondingauthor{Timothy Hospedales}{t.hospedales@ed.ac.uk}

\icmlkeywords{Machine Learning, ICML}

\vskip 0.3in
]



\printAffiliationsAndNotice{} 

\begin{abstract}
Adversarial examples are of wide concern due to their impact on the reliability of contemporary machine learning systems. Effective adversarial examples are mostly found via white-box attacks. However, in some cases they can be transferred across models, thus enabling them to attack black-box models. In this work we evaluate the transferability of three adversarial attacks - the Fast Gradient Sign Method, the Basic Iterative Method, and the Carlini \& Wagner method, across two classes of models - the VGG class(using VGG16, VGG19 and an ensemble of VGG16 and VGG19), and the Inception class(Inception V3, Xception, Inception Resnet V2, and an ensemble of the three). We also outline the problems with the assessment of transferability in the current body of research and attempt to amend them by picking specific "strong" parameters for the attacks, and by using a L-Infinity clipping technique and the SSIM metric for the final evaluation of the attack transferability.
\end{abstract}

\section{Introduction}

Image classification is the task of assigning an image a specific class based on the image contents. Convolutional neural networks (CNNs) \cite{LeCun:1999:ORG:646469.691875} are state-of-the-art neural networks widely used for this purpose. However,  neural networks are vulnerable to adversarial examples\cite{DBLP:journals/corr/SzegedyZSBEGF13}. \\ \\
Adversarial examples are inputs created with the purpose of fooling a machine learning model. In the context of image classification, this means distorting the pixel values of an initially correctly classified image until misclassification. \\ \\
Adversarial examples pose challenging questions, such as how neural networks and humans differ in classifying images. Adding noise to an image does not usually deceive a human annotator, however, neural networks can be easily fooled in many cases. \\ \\
From a security perspective, adversarial examples are a potential risk when introduced to a variety of machine learning applications, such as self-driving cars, drone control, facial and speech recognition systems and more. For example, \cite{DBLP:journals/corr/abs-1804-05296} elaborates on how bad actors in the heavily funded healthcare sector are able to take advantage of adversarial attacks for misdiagnoses, leading to unnecessary and expensive medical procedures. \\ \\
Adversarial images are created via adversarial attacks. Adversarial attacks target a specific model, and look to change the input of this model until misclassification. In a "white-box" setting, the attacker has access to the architecture and parameters of the model. Adding random noise without taking advantage of the available knowledge of the model will produce adversarial inputs of much lower quality. \\ \\
Knowing the parameters and/or architecture of a model, however, is not always possible. A malicious agent attacking a web service, for example, will not typically know the inner details of the algorithm employed by the service. In this "black-box" setting, the agent may attempt to attack the service by running an adversarial attack on either a custom local model, or on a publicly available model, in a white-box way, and then sending the adversarially created input over to the remote service. An example of this is the attack on Clarifai.com \cite{DBLP:journals/corr/LiuCLS16}. The property of an adversarial example created by one system with known architecture and parameters, to transfer to another unknown black-box system, is called transferability. \\ \\
In this work we focus on the "one-shot" black-box transferability of adversarial attacks. By this we mean that an attacker does not receive feedback from the black-box system, and therefore cannot make any further modifications of the adversarial image depending on this feedback. More specifically, we attempt to measure the transferability of three different adversarial attacks for risk assessment across five standard image classification models pre-trained on the ImageNet dataset: VGG16, VGG19, Inception V3, Xception and Inception Resnet V2, and across two ensembles. Do small differences in the model architecture account for small differences in the transferability? Since the models are well-known, highly-performant, free, and publicly available, web services have an incentive to adapt them to the services' specific use-cases via transfer learning. This could lead to overuse of the models in the public space and motivates the choice of models explored in this work. \\ \\
In the context of previous work, research has been done on our explored standard models to evaluate the effectiveness or transferability of adversarial attacks \cite{DBLP:journals/corr/abs-1808-01688, DBLP:journals/corr/LiuCLS16}.  However, this is usually done only for single or multiple parameter settings of the attacks. The parameters are usually chosen, so that the visual change in the attacked images is imperceptible. This "parameter overfitting" may lead to unfair or inexhaustive comparisons due to insufficient values of the parameters being tested and the varying importances and roles of the parameters used in the different algorithms. This gives motivation for using a different way of assessing transferability(sec. \ref{sec:linifinity}) and for further exploration into the problem of assessing how attacks can be calibrated in order to draw consistent and fair conclusions. \\

\section{Background}

\subsection{Adversarial Attacks Frameworks}
Two out-of-the-box adversarial attacks frameworks were tested in this work. The versions of the frameworks with the necessary modifications used for the experiments can be found \href{https://github.com/deqncho2/foolbox}{{\textbf{here}\footnote{https://github.com/deqncho2/foolbox}}} and \href{https://github.com/deqncho2/art}{{\textbf{here}\footnote{https://github.com/deqncho2/art}}}.

\subsubsection{Adversarial Robustness Toolbox}
The Adversarial Robustness Toolbox (ART) \cite{DBLP:journals/corr/abs-1807-01069} is a tool developed by IBM supporting various attacking and defensive techniques, such as adversarial training, defensive distillation \cite{DBLP:journals/corr/PapernotMWJS15}, and denoisers. The library was used in its raw format for the experiments.
\subsubsection{Foolbox}
Foolbox is library which also implements a variety of attacks such as informed noise attacks, gradient based attacks, and attacks optimizing different norms. This package was slightly modified for the experiment setup.

\subsection{Standard Image Classification Networks}

The following models were the default Keras implementations pre-trained on the ImageNet dataset.

\subsubsection{VGG Family}
The first evaluated family of models is the VGG family. The VGG16 architecture was first introduced in \cite{2014arXiv1409.1556S}. For accurate classification, the networks expect a RGB image to be preprocessed by flipping the R and B dimensions and subtracting 103.939 from the blue, 116.779 from the green, and 123.68 from the red channel. 
\paragraph{VGG16}
The VGG16 implementation in Keras consists of five blocks. The first two blocks use two consecutive convolutional layers of filter size 3, and \textit{same} padding. The last three blocks have the same specifics, but use three convolutions instead of two. The activation function is \textit{Relu}. The number of convolutional filters doubles each block. After the convolutions there is a max pooling layer of size and stride 2. At the end of the blocks there are two fully connected layers with Relu activation of size 4096 and finally a Softmax layer for the 1000 ImageNet categories. 

The network has achieved 7.3\% top-5 error rate on the 2014 ImageNet challenge.
\paragraph{VGG19}
The VGG19 architecture has the same architecture as VGG16, but instead of three, it has four convolutional layers in the last three blocks. It allows for deeper feature representations than VGG16, but does not yield much more accurate results than VGG16 on the ImageNet challenge.

An ensemble of the two models was also evaluated against the attacks.
\begin{figure}[H]
    \centering
    \includegraphics[height=45mm]{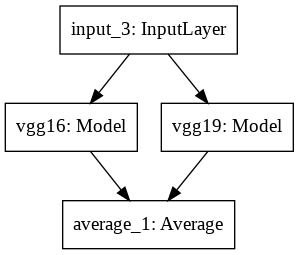}
    \caption{VGG Ensemble}
\end{figure}

\subsubsection{Inception Family}

GoogLeNet, also known as Inception-V1 was first introduced in 2014 \cite{DBLP:journals/corr/SzegedyLJSRAEVR14} and exploited the idea of an "inception modules"\cite{DBLP:journals/corr/LinCY13}. The inception modules are feature maps combining contexts of different sizes to obtain different types of patterns, which removes the need for manual selection of filter sizes and reduces computational and memory cost via convolution dimensionality reductions. 

The GoogLeNet architecture combines these modules and has 6.7\% error rate on the 2014 ImageNet challenge and outperforms VGG16 despite being a lot smaller(55 MB vs 490 MB).

The inception models' preprocessing require the pixels of an image to be divided by 255, subtracted by 2, and multiplied by 2, so the original [0-255] pixel range is mapped to [-1-1].

\paragraph{Inception V3}
Inception V2 and V3 were introduced in the same paper\cite{DBLP:journals/corr/SzegedyVISW15}. In it, several improvements were done to the original Inception architecture. 5x5 convolutions were factorized to two 3x3 convolutions, which sped up computation. Furthermore, the authors found that deeper representations lose information due to loss of dimensions and thus information. Therefore, they made the inception modules wider to keep the number of channels high. For version 3, RMSProp Optimizer and batch normalization\cite{DBLP:journals/corr/IoffeS15} were added, 7x7 convolutions were factorized, and label smoothing was done on the class predictions to prevent the network from growing too confident and to avoid overfitting.
\paragraph{Xception} Xception was introduced in \cite{DBLP:journals/corr/Chollet16a} and replaces the Inception network convolutions with depthwise separable convolutions. It has roughly the same number of parameters as Inception-V3, but slightly outperforms it on the ImageNet dataset. It is based on the strong assumption that spatial and depthwise correlations can be decoupled, leading to the name of the network Xception (from "Extreme Inception"). After performing a 1x1 convolution on a feature map, it uses \textit{n} convolutional filters, where \textit{n} is the number of channels of the input, and each filter iterates only over the specific channel, instead of using filters that iterate over all channels. The network also makes use of residual blocks and takes into account previous feature maps.
Xception has slightly smaller memory requirements than Inception V3.

\paragraph{Inception Resnet V2}

Our final model is Inception Resnet V2. It incorporates elements of Residual Networks \cite{DBLP:journals/corr/HeZRS15} into the Inception V4 architecture. It has the same architecture as Inception Resnet V1 with the exception of the "stem", but has different hyperparameter settings. Both versions were introduced in the same paper \cite{DBLP:journals/corr/SzegedyIV16}. Inception V4 and Inception Resnet V2 have similar computational complexity despite the latter being deeper. They both utilize reduction blocks which reduce the size of the output, and Block A is the same in both architectures. Inception V4 and Inception Resnet V2 also differ by the types of modules they use. The Inception Resnet versions incorporate residual connections in their modules.

For network stability, activation scaling is used for residual layers deeper in the architecture.

Out of the three models, Inception Resnet V2 is the deepest and the differs the most from the other two. It requires twice the memory and computational operations as compared to Inception V3. However, it outperforms Inception V3 on the ILSVRC 2012 image classification benchmark with a top-5 accuracy of 95.3\%(with 93.9\% for Inception V3) and top-1 accuracy of 80.4\%(compared to 78.0\%). 

Finally, as before, the concatenation of these models was also evaluated by averaging their predictions(using the argument $use\_logits=True$ in ART and $predicts = "logits"$ in Foolbox).

\begin{figure}
    \centering
    \includegraphics[width=\columnwidth]{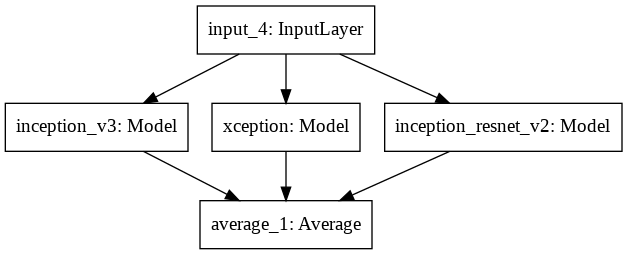}
    \caption{Inception Ensemble.}
\end{figure}

\subsection{Adversarial Attacks and Defense}

Statistics were gathered for three attacks - the Fast Gradient Sign Method, the Basic Iterative Method,and the Carlini Wagner method.
\paragraph{Fast Gradient Sign Attack Method}
The Fast Gradient Sign Method was first introduced in \cite{Goodfellow2015ExplainingAH} and exploits the linear nature of neural networks. It is an attack which adds noise to the original input by taking the sign of the gradient of the loss function \textit{J} of a trained model with respect to the inputs, and adding it to the original input. A distortion parameter $\epsilon$ controls how much the input is perturbed. The attack is defined by the equation:

\[ x^{adv} = x + \epsilon*\textit{sign} (\nabla_x J(x, y_{true})) \]
where $x^{adv}$ is the calculated adversarial image, $x$ is the original image, $y_{true}$ is the true label of the image, and $\nabla_x J$ is the Jacobian of the loss function with respect to the image.

It is a fast and reliable method to find adversarial examples, although not the most widely used for finding realistic looking adversarials, as high-success rate attacks yield more noticeable perturbations than other more advanced methods.
\paragraph{Iterative Fast Gradient Sign Attack Method}
The Iterative Fast Gradient Sign Method takes \textit{T} gradient steps of magnitude $\alpha=\epsilon/T$ instead of a single step. It is also called Basic Iterative Method \cite{DBLP:journals/corr/KurakinGB16} in ART and does not apply clipping after each iteration(however, there is an argument for that, which is the algorithm described in \cite{DBLP:journals/corr/KurakinGB16}). The equations are:
\[x_0^{adv} = x\]
\[x_{t+1}^{adv} = x_{t}^{adv} + \alpha*\textit{sign}(\nabla_x J(x_{t}^{adv}, y_{true}))\]
where $x_{t}^{adv}$ is the adversarial found at iteration $t$.

\paragraph{Carlini and Wagner Attack Method}
The Carlini and Wagner \cite{DBLP:journals/corr/CarliniW16a} method is currently one of the strongest adversarial attacks. It was used to break defensive distillation(a method of defense that was able to reduce the success rate of previous attacks' ability to find adversarial examples from 95\% to 0.5\%). It achieved success rate of 100\% on both distilled and undistilled networks for all three norms of the attack.
The attack solves the optimization problem of minimizing the distance \textit{D} between the original image and the adversarial:

\begin{mini}{}{D(x, x + \mu)} {}{}
\addConstraint{f(x+\mu)}{\leq 0}
\end{mini}

The optimization function \textit{f} is \[ f(x+\mu) = max(max\{Z(x+\mu)_i : i \neq t\} - Z(x+\mu)_t, -k) \]
where Z is the softmax output for the most probable class different from the targeted class.

The \textit{k} parameter controls the confidence with which the misclassification occurs. The attack uses the Adam optimizer which allows for the tuning of the learning rate. Several other parameters are available in the frameworks, which did not seem to affect the attack success as much as the learning rate(most contributory) and the confidence constant(second most contributory). 


\section{Experiments and Evaluation}
\label{sec:linifinity}
A large body of research has so far concentrated on finding attacks and parameters with high success rate for low perturbations. These evaluations may sometimes find adversarial examples within one pixel \textit{L-infinity} distance from the original image(i.e. each pixel of the adversarial is within a value of one of the original image). Many papers compare attacks by, for example, listing a small number of \textit{L2/L-infinity} norm thresholds for the perturbations and reporting accuracy percentages. \\

Though useful for evaluation of minimal perturbations, these comparisons are not universally practical due to the wide-spread image formats using integer representation. Converting an adversarially found numpy floating point array to an image is not possible, except if very specific and not widely spread floating point image formats, such as \textit{.tiff} are used. These formats provide high-quality images  and are usually employed in niche domains(e.g. X-ray imaging). A majority of common formats such as the \textit{.png} and \textit{.jpg} format use 8-bit integer representation for their color channels and pixel values[0-255]. Many papers take this into account and measure minimal perturbations from images rounded to integer values, though some do not mention it explicitly. \\

This means that until floating point image formats are widely used, an attacker has to round the pixel values, which could potentially overturn the adversarial training process and make the prediction of the white-box-attacked classifier correct again, which makes evaluation of minimal perturbations not valuable in some cases. The effects of rounding the values may vary. For this work, metrics were gathered on valid images that were mapped back to their original [0-255] ranges.
In order to examine transferability fairly, a fair metric is needed. An attack might be more transferable than another just because it changes the image more, not because the changes are necessarily better or "stronger". An initial idea for a fair metric was to pick several thresholds for some \textit{L-norm} distance from the original image and compare the transferability across these thresholds. \\

This is the method used by many works in the literature. However, a problem with this is that the ranges of perturbations found by different attacks can vary by orders of magnitude, which makes picking appropriate thresholds to evaluate transferability problematic. Table \ref{tab:a} shows normalized \textit{MSE} distances for iterations from runs for some of the different tested attacking algorithms for a sample image using the Foolbox. It can be seen that different attacks start finding adversarials at different distances(first entries of the table), some start to converge, and others take different-sized steps.
\begin{table*}[h]
\centering
\begin{tabular}{|l|l|l|l|l|l|}
\hline
\textbf{Saliency Map}  & \textbf{Blended Noise}  & \textbf{DeepFool} & \textbf{Iterative Gradient}  & \textbf{Contrast Reduction}                       \\ \hline
6.07e-04 & 3.49e-02 & 7.96e-06 & 1.09e-03  & 9.48e-02                      \\ \hline
6.16e-04  & 3.50e-02 & 1.02e-05 & 1.24e-03 & 9.50e-02   \\ \hline
6.20e-04 & 3.52e-02 & 1.04e-05 & 1.40e-03 & 9.52e-02  \\ \hline
6.21e-04 & 3.54e-02 & 1.18e-05 & 1.54e-03  & 9.54e-02                       \\ \hline
6.22e-04  & 3.55e-02 & 1.19e-05 & 1.73e-03 & 9.56e-02   \\ \hline
6.25e-04 & 3.57e-02 & 1.19e-05 & 1.87e-03 & 9.58e-02  \\ \hline
6.29e-04 & 3.58e-02 & 1.19e-05 & 2.05e-03  & 9.60e-02                       \\ \hline
6.30e-04  & 3.60e-02 & 1.19e-05 & 2.22e-03 & 9.62e-02   \\ \hline
6.34e-04 & 3.62e-02 & 1.19e-05 & 2.40e-03 & 9.63e-02  \\ \hline
6.39e-04 & 3.63e-02 & 1.19e-05 & 2.57e-03  & 9.64e-02                       \\ \hline
6.44e-04  & 3.65e-02 & 1.19e-05 & 2.77e-03 & 9.66e-02   \\ \hline
6.49e-04 & 3.67e-02 & 1.19e-05 & 2.96e-03 & 9.68e-02  \\ \hline
6.54e-04 & 3.68e-02 & 1.19e-05 & 3.16e-03  & 9.69e-02                       \\ \hline
...  & ... & ... & ... & ...   \\ \hline
6.58e-02 & 9.43e-02 & 1.19e-05 & 9.62e-02 & 9.70e-02  \\ \hline
6.58e-02 & 9.46e-02 & 1.19e-05 & 9.63e-02  & 9.72e-02                       \\ \hline
6.58e-02  & 9.48e-02 & 1.19e-05 & 9.64e-02 & 9.74e-02   \\ \hline
6.58e-02 & 9.51e-02 & 1.19e-05 & 9.65e-02 & 9.76e-02  \\ \hline
6.59e-02 & 9.54e-02 & 1.19e-05 & 9.67e-02  & 9.78e-02                       \\ \hline
6.59e-02  & 9.56e-02 & 1.19e-05 & 9.68e-02 & 9.80e-02   \\ \hline
6.59e-02 & 9.59e-02 & 1.19e-05 & 9.70e-02 & 9.82e-02  \\ \hline
6.59e-02 & 9.62e-02 & 1.19e-05 & 9.70e-02  & 9.84e-02                      \\ \hline
6.59e-02  & 9.64e-02 & 1.19e-05 & 9.72e-02 & 9.86e-02   \\ \hline
Interrupted & Interrupted & Interrupted & Interrupted & Interrupted  \\ \hline
\end{tabular}
\caption{Normalized \textit{MSE} for adversarials across iterations}
\label{tab:a}
\end{table*}
A way to amend this problem is to restrict the comparison to a family of similar algorithms which modify the image in comparable ranges. However, this method is too restrictive in general as new attacks emerge, and thresholds for existing attacks are difficult to choose. \\

Another method is to let an attack run its course and then post-process the adversarial image by clipping it to the closest point on a ball of a given radius away from the original image. This method is how the experiments were set up.
For the experiments, a dataset of 496 images, all classified correctly with 0.98 confidence or more than all five classifiers, was composed. Images as far away from the decision boundaries were preferred in order to discourage the effectiveness of random attacks. \\

The images were "trained" for the seven classifiers in a white-box way using an untargeted attack(the attack changes the image to any other class instead of a specific target class). Then, each of the models was attacked in a black-box manner by simply taking the accuracy of the predictions of the classifiers for the clipped trained images. \\

The experiment setup and technical details are as follows. 
The dataset was created by gathering five random classes of images with a total of 497 images (about 100 images of each class) which were passed through the VGG or Inception preprocessing functions. They were center-cropped to have dimensions of 224x224x3 and were ensured to have correct and confident predictions for all five models after the preprocessing(during the course of this work, the Keras version was updated to not allow 224x224 images for the Inception models, but only 299x299. Older Keras versions work for this, e.g. 2.1.5). \\

For the training stage, the images were passed through the preprocessing functions for the VGG or Inception models. Then, the attacking algorithms were run, and were left to "wonder around". After, reverse preprocessing was performed, the images were clipped to a minimum pixel value of 0, and maximum of 255, in case they were out of the allowed range after the reverse processing operation(as the API of the attacks did not allow for different color channels to have different minimum and maximum values, which is the case for the VGG preprocessed \textit{numpy} arrays). Then, the images were clipped again to different \textit{L-infinity} norm ranges, as described below. Finally, the images were rounded to the nearest integer and results are gathered. The clipping procedure used was \textit{numpy}'s function "clip". \\

Using this methodology a strong attack which fools most of the images in the source model had to be found right after the preprocessing steps. Parameter search for the attacks was done manually, due to slow training times that discouraged standard parameter search procedures. \\

Different \textit{L-infinity} clipping ranges - 0, 5, 10, 15, until 150 (each pixel value of the adversarial must be within a maximum of each of these values from the original image, consult figs. \ref{fig: appendix}, \ref{fig: appendix2} and \ref{fig: appendix3} for images of different noise levels) were used in the presented results.

\subsection{Attack Parameters}
In this section we present the parameters and the toolboxes used for the experiments. Parameters other than the mentioned ones were left to their default values. Table \ref{tab:acc} shows the different classifiers' accuracies in percentages right after the preprocessing and attack end, and before the postprocessing operations described in the section \ref{sec:linifinity}.
\subsubsection{FGSM Attack - ART}

For the \textit{L-Infinity} FGSM implementation, ART was used with the parameter $\epsilon=100$.

\subsubsection{I-FGSM Attack - ART}
The parameters used were a step size of 1, and number of iterations of 100 to match FGSM's distortion. The attack optimizes \textit{L-Infinity} as well.

\subsubsection{Carlini \& Wagner Attack - Foolbox}
For this \textit{L2-norm} version of the attack, Foolbox was used with parameters $\textit{max\ iterations}=100$, $\textit{learning rate}=7$, and $\textit{confidence}=50$. The rest of the parameters were left to default at $\textit{initial const}=1e-2$ and $\textit{binary search steps}=5$.

\begin{table}[H]
\begin{tabular}{|l|l|l|l|l|l|}
\hline
\textbf{Classifier} & \textbf{FGSM} & \multicolumn{2}{l|}{\textbf{I-FGSM}} & \multicolumn{2}{l|}{\textbf{C\&W}} \\ \hline
VGG16               & 0.0\%         & \multicolumn{2}{l|}{1.609\%}         & \multicolumn{2}{l|}{0.0\%}         \\ \hline
VGG19               & 0.6\%         & \multicolumn{2}{l|}{2.01\%}          & \multicolumn{2}{l|}{0.0\%}         \\ \hline
Inception V3        & 0.0\%         & \multicolumn{2}{l|}{0.0\%}           & \multicolumn{2}{l|}{0.0\%}         \\ \hline
Xception            & 0.0\%         & \multicolumn{2}{l|}{0.0\%}           & \multicolumn{2}{l|}{0.0\%}         \\ \hline
Inception Resnet 2  & 0.0\%         & \multicolumn{2}{l|}{0.0\%}           & \multicolumn{2}{l|}{0.0\%}         \\ \hline
VGG Ensemble        & 0.0\%         & \multicolumn{2}{l|}{1.207\%}         & \multicolumn{2}{l|}{0.0\%}         \\ \hline
Inception Ensemble  & 0.0\%         & \multicolumn{2}{l|}{0.0\%}           & \multicolumn{2}{l|}{0.603\%}       \\ \hline
\end{tabular}
\caption{Accuracy of the classifiers under different attacks before postprocessing (smaller numbers indicate a more successful attack)}
\label{tab:acc}
\end{table}

\subsection{Visual Metrics Motivation}
The results(after postprocessing) when using \textit{L-Infinity} as a metric for visual perturbation are shown in fig. \ref{fig: linf}. However, since the L-Infinity metric is not indicative of noticeable visual perturbation(for example, an adversarial which has 1 pixel distorted by a value of 100 is the same as an adversarial which has all of its pixels distorted by a value of 100 under the \textit{L-Infinity} metric), we aim to gather results more correlated with human perception of noise levels. Therefore, visual metrics other than \textit{L-Infinity} were considered when reorganizing the results for the \textit{L-Infinity}-clipped images.

\subsection{Visual Metrics}
Three other metrics were considered to reorganise the \textit{L-Infinity} results in terms of metrics which are more correlated with how human perceive images.

\afterpage{
\begin{figure*}[p]
\centering
\subfigure[VGG16]
{\includegraphics[width=.49\columnwidth]{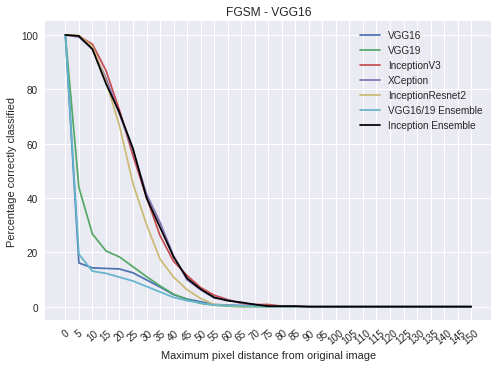}}
\subfigure[VGG19]
{\includegraphics[width=.49\columnwidth]{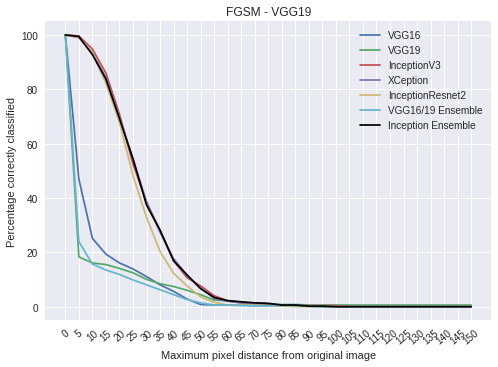}}
\subfigure[Inception V3]
{\includegraphics[width=.49\columnwidth]{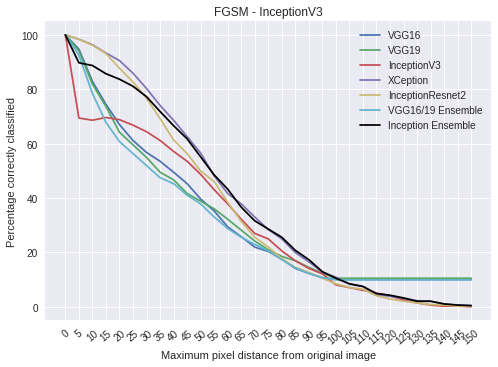}}
\subfigure[Xception]
{\includegraphics[width=.49\columnwidth]{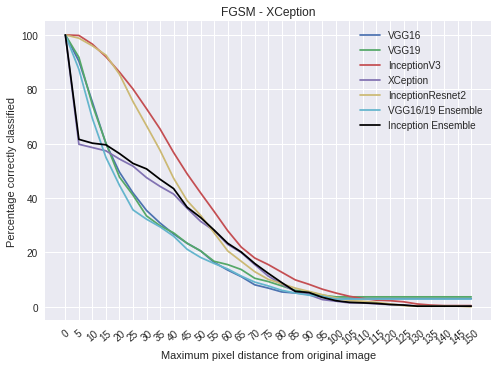}}
\vskip -0.1in
\centering
\subfigure[Inception Resnet 2]
{\includegraphics[width=.55\columnwidth]{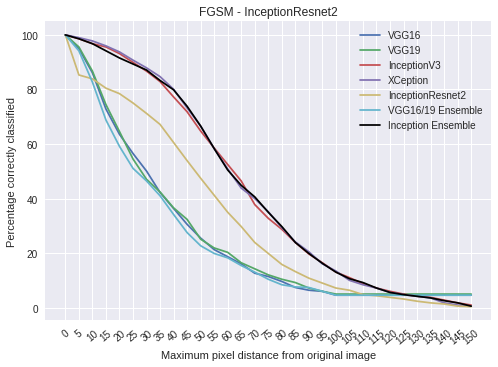}}
\subfigure[VGG Ensemble]
{\includegraphics[width=.55\columnwidth]{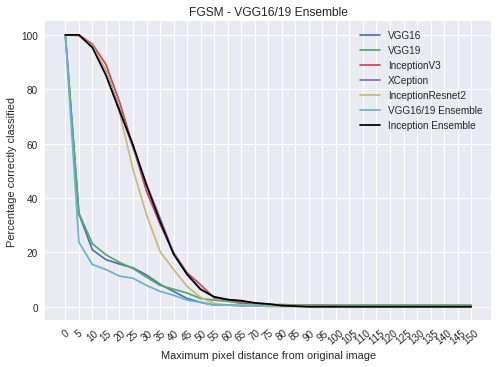}}
\subfigure[Inception Ensemble]
{\includegraphics[width=.55\columnwidth]{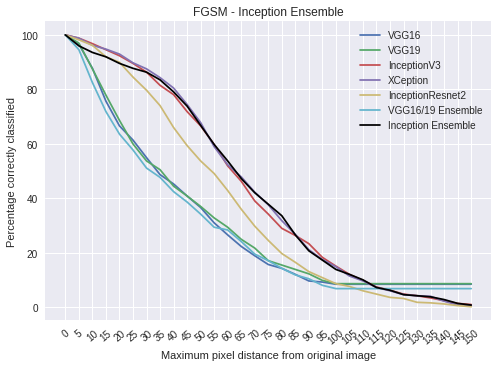}}
\subfigure[VGG16]
{\includegraphics[width=.49\columnwidth]{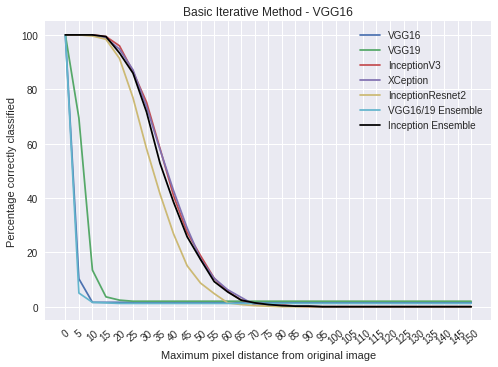}}
\subfigure[VGG19]
{\includegraphics[width=.49\columnwidth]{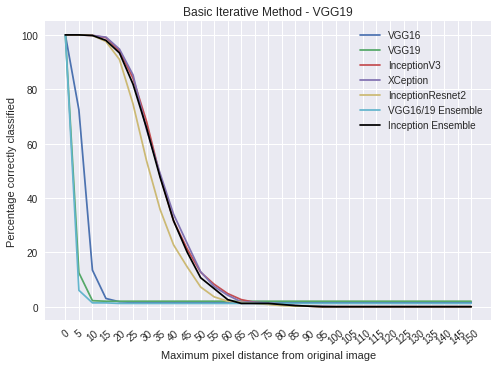}}
\subfigure[Inception V3]
{\includegraphics[width=.49\columnwidth]{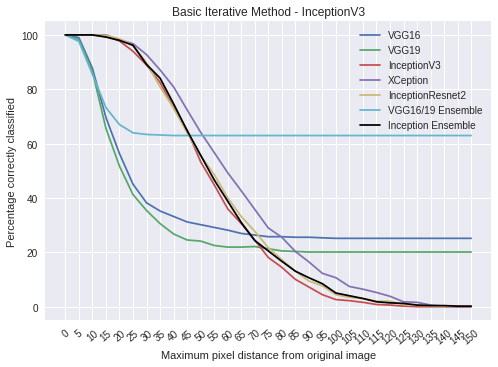}}
\subfigure[Xception]
{\includegraphics[width=.49\columnwidth]{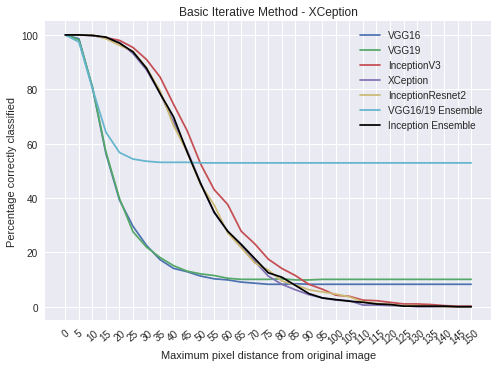}}
\vskip -0.1in
\centering
\subfigure[Inception Resnet 2]
{\includegraphics[width=.55\columnwidth]{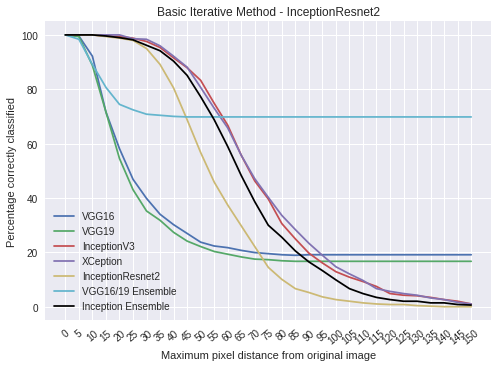}}
\subfigure[VGG Ensemble]
{\includegraphics[width=.55\columnwidth]{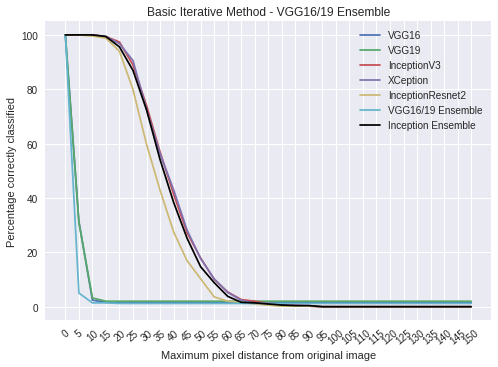}}
\subfigure[Inception Ensemble]
{\includegraphics[width=.55\columnwidth]{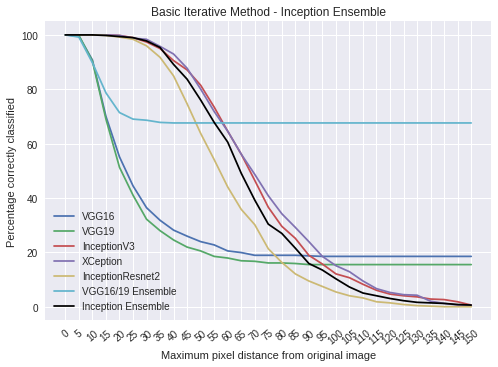}}
\subfigure[VGG16]
{\includegraphics[width=.49\columnwidth]{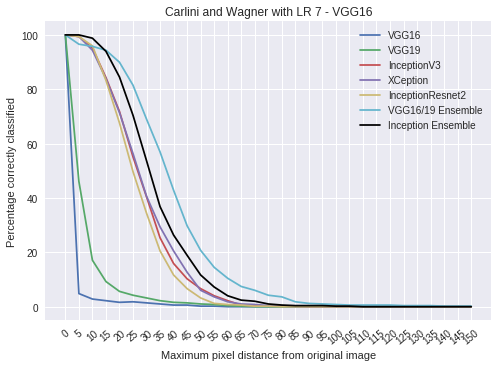}}
\subfigure[VGG19]
{\includegraphics[width=.49\columnwidth]{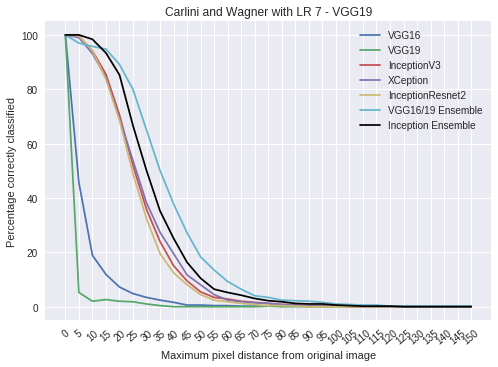}}
\subfigure[Inception V3]
{\includegraphics[width=.49\columnwidth]{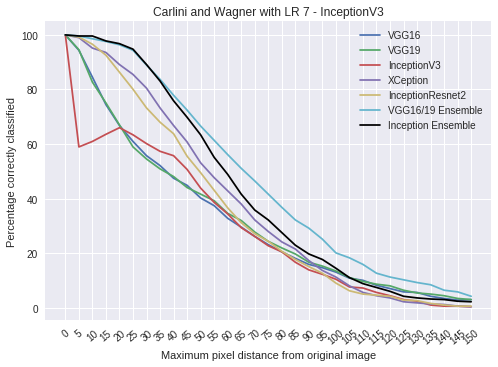}}
\subfigure[Xception]
{\includegraphics[width=.49\columnwidth]{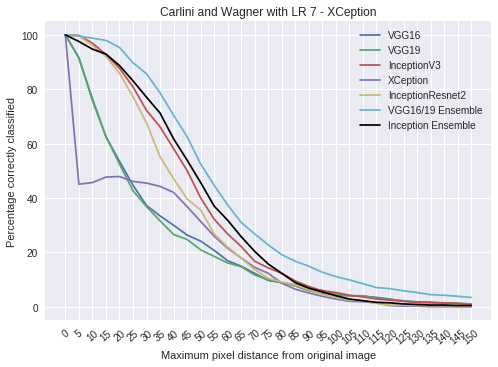}}
\vskip -0.1in
\centering
\subfigure[Inception Resnet 2]
{\includegraphics[width=.55\columnwidth]{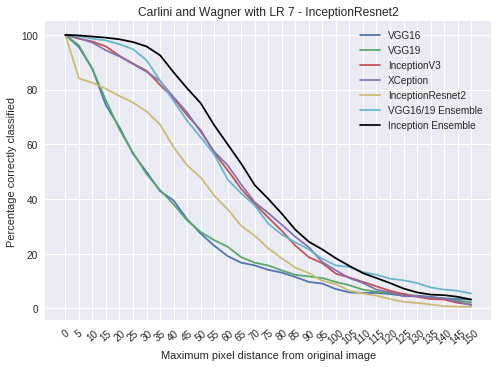}}
\subfigure[VGG Ensemble]
{\includegraphics[width=.55\columnwidth]{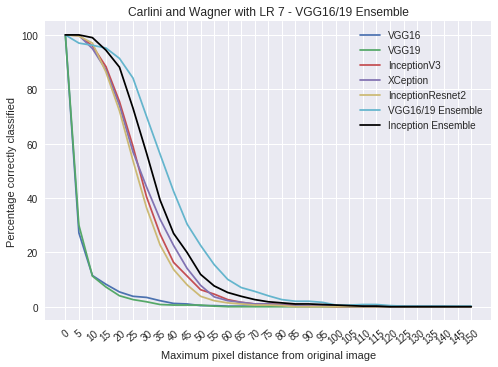}}
\subfigure[Inception Ensemble]
{\includegraphics[width=.55\columnwidth]{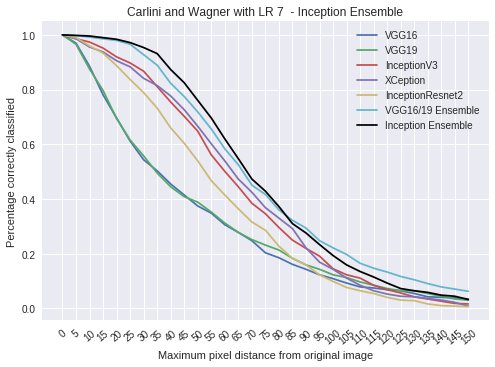}}
\vskip -0.1in
\caption{Performance of every classifier when attacked by each of the 7 classifiers, for all 3 attacks, under the L-Infinity metric. For example, (a) shows the defensive accuracy of VGG16 when attacked by clipped adversarial images created by all 7 classifiers, using the FGSM attack(title of plot signifies the defending classifier and the used attack).}
\label{fig: linf}
\end{figure*}
\clearpage
}

\subsubsection{Inception Score}
The first tested visual metric was the Inception Score (IS), first introduced in \cite{DBLP:journals/corr/SalimansGZCRC16}. It serves as a measure for the performance of Generative Adversarial Networks(GANs). The \textit{IS} takes a list of images as input and returns a score. The \textit{IS} was calculated using the implementation available \href{https://github.com/tsc2017/Inception-Score}{{\textbf{here}\footnote{https://github.com/tsc2017/Inception-Score}}}, where the lists of the clipped images of the different various \textit{L-Infinity} clipping ranges were passed through the scoring function. After examination of the images, the perturbation for the clipping ranges appeared to be gradually increasing. 

However, there was no pattern to the \textit{IS}(fig. \ref{fig:2}). Therefore the metric was not chosen to reorganize the results. The \textit{IS} is not typically used in the literature to measure the levels of adversarial noise. We observe that this metric is ineffective for this purpose.

\subsubsection{Mean Absolute Distance (MAD)}
Commonly used metrics in the literature to measure minimum perturbations of adversarial examples include the Mean Squarred Distance(MSD) and Mean Absolute Distance(MAD). \textit{MAD} was chosen as another potential metric for reorganization of the results (fig. \ref{fig:2}) and measures the mean absolute difference between the clipped images of the different maximum ranges and the originals.
\subsubsection{Structural Similarity Index (SSIM)}
\begin{figure}[H]
\centering
\includegraphics[width=0.15\textwidth]{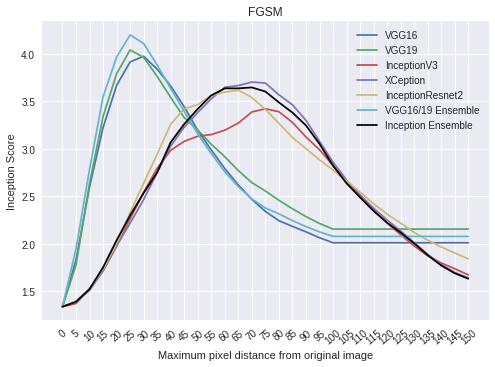}~
\includegraphics[width=0.15\textwidth]{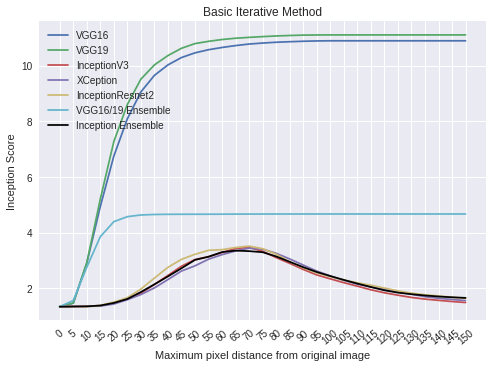}~
\includegraphics[width=0.15\textwidth]{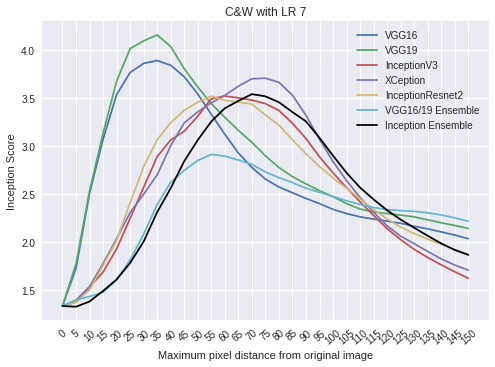}
\includegraphics[width=0.15\textwidth]{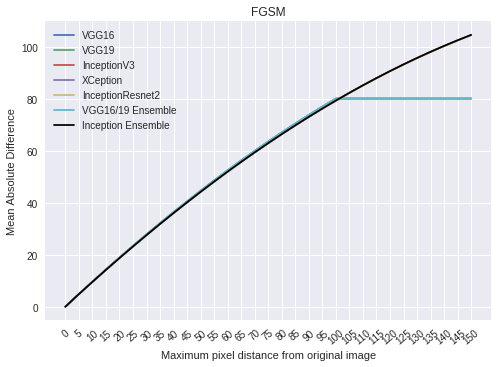}~
\includegraphics[width=0.15\textwidth]{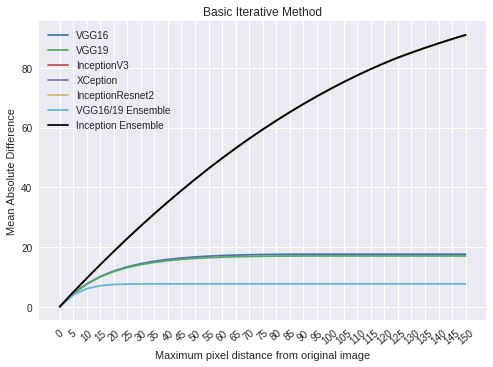}~
\includegraphics[width=0.15\textwidth]{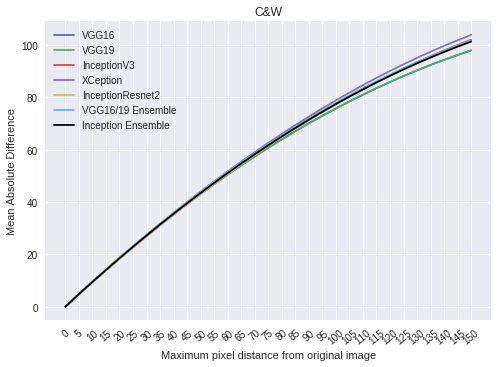}
\includegraphics[width=0.15\textwidth]{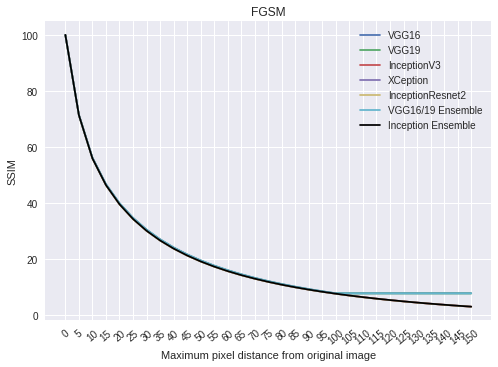}~
\includegraphics[width=0.15\textwidth]{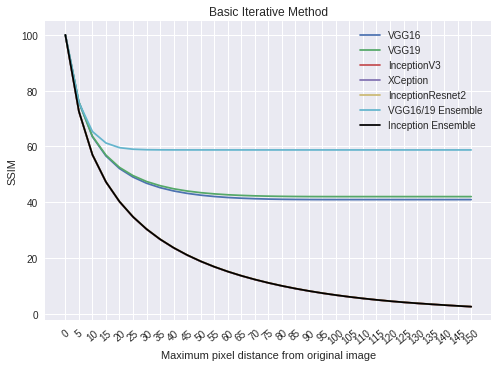}~
\includegraphics[width=0.15\textwidth]{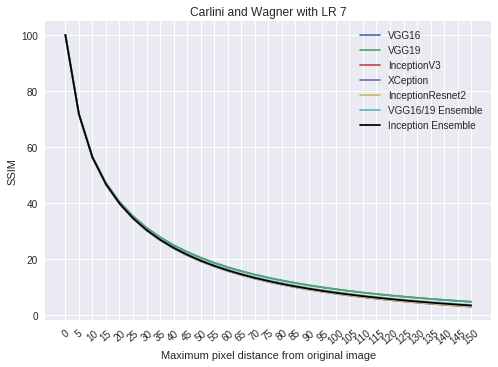}

\caption{The three visual metrics considered for the reorganization of results - Inception Score of the adversarials of varying clip strengths(top), Average Mean Absolute Distance of adversarials of varying clip strengths from the original images(middle), and Average Structural Similarity Index of adversarials of varying clip strengths from the originals(bottom).} 
\label{fig:2}
\end{figure}

The final metric tested was the \textit{SSIM}(fig. \ref{fig:2}) and can be used to measure the quality degradation between 2 images, commonly reported from 1(no degradation) to 0 or from 100 to 0. This measure has been mentioned in the image analysis literature to correlate more with human perception than \textit{MAD}\cite{article}. 

Therefore, it was chosen for reorganization of the results(fig. \ref{fig:ssimreorg}). \href{http://scikit-image.org/docs/dev/auto_examples/transform/plot_ssim.html}{{\textbf{Scikit-image}} } was used for the calculation of the \textit{SSIM}. For the different clipping ranges, the mean \textit{SSIM} between the adversarial images, and their corresponding originals was calculated. 

\subsection{Discussion}
\label{sec:discussion}
\begin{figure}[hbt]
\centering
\begin{tabular}{cc}
  \includegraphics[scale=0.3]{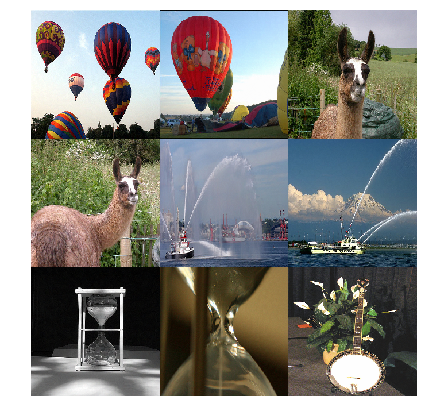} &   \includegraphics[scale=0.3]{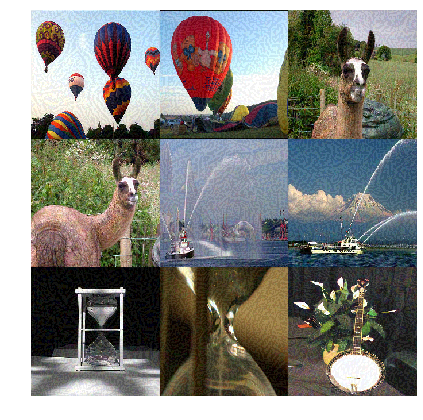} \\
 Original images & Clip range - 10, avg. transferability - 49\% \\[6pt]
 \\[6pt]
  \includegraphics[scale=0.3]{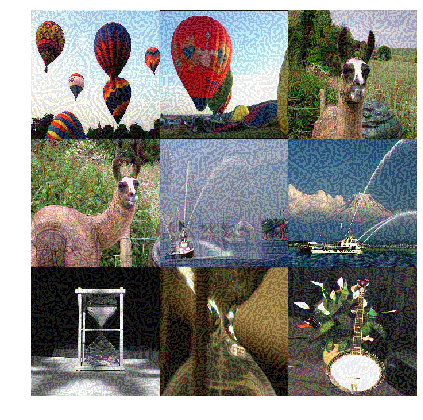} &   \includegraphics[scale=0.3]{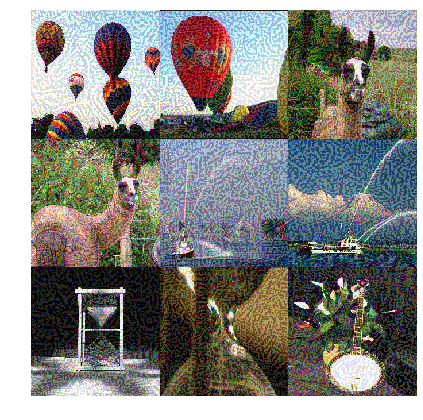} \\
 Clip range - 20, avg. transferability - 63\% & Clip range - 30, avg. transferability - 71\% \\[6pt]
 \\[6pt]
  \includegraphics[scale=0.3]{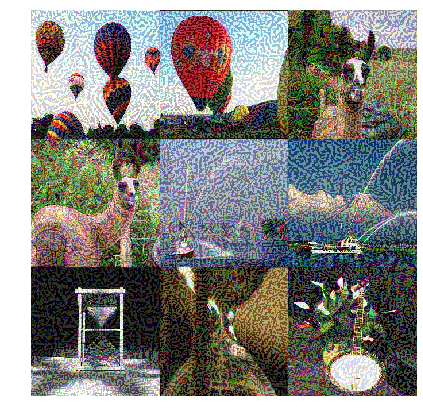} &   \includegraphics[scale=0.3]{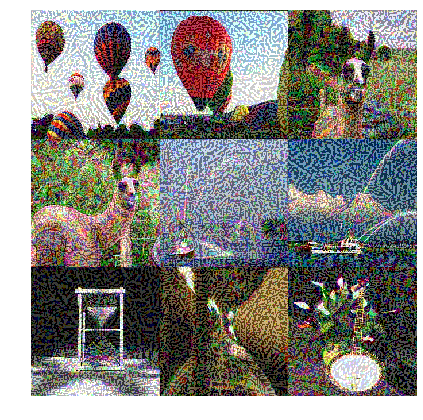} \\
 Clip range - 40, avg. transferability - 77\% & Clip range - 50, avg. transferability - 83\% \\[6pt]
 \\[6pt]
  \includegraphics[scale=0.3]{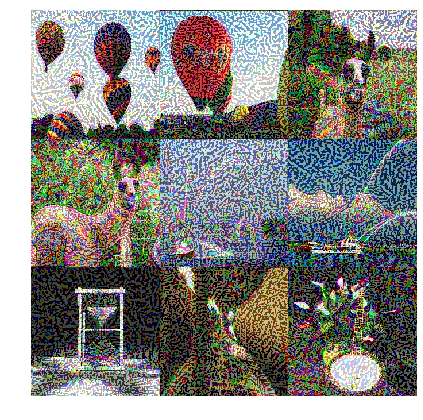} &   \includegraphics[scale=0.3]{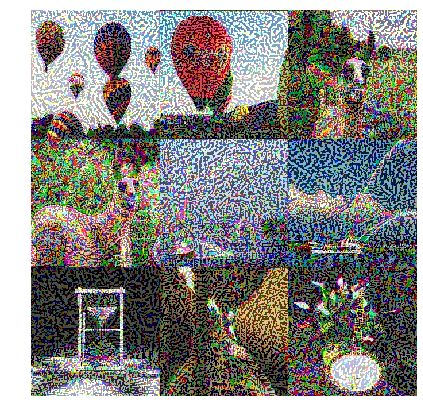} \\
Clip range - 60, avg. transferability - 87\% & Clip range - 70, avg. transferability - 90\% \\[6pt]
 \\[6pt]
\end{tabular}
\caption{Images for the FGSM attack trained on the VGG-Ensemble of different \textit{L-Infinity} clipping ranges and the average transferability(rounded) of these ranges across all models, including the source}
\label{fig: appendix}
\end{figure}

\begin{figure}
\centering
\begin{tabular}{cc}
  \includegraphics[scale=0.3]{proj/images/orig.png} &   \includegraphics[scale=0.3]{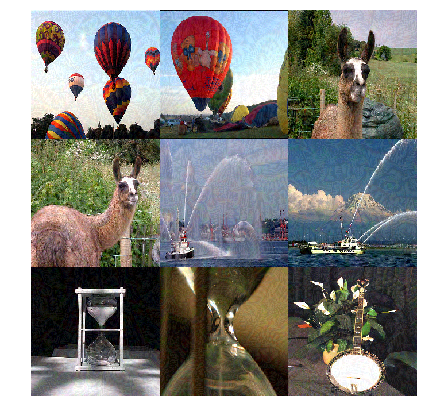} \\
 Original images & Clip range
 - 10, avg. transferability - 48\% \\[6pt]
 \\[6pt]
  \includegraphics[scale=0.3]{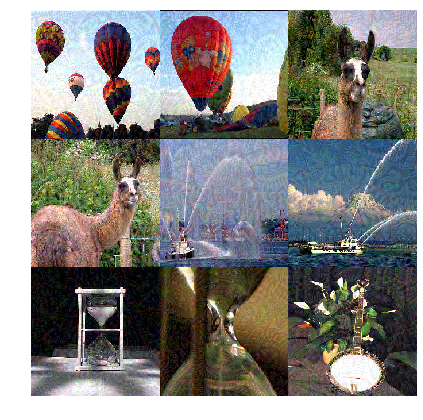} &   \includegraphics[scale=0.3]{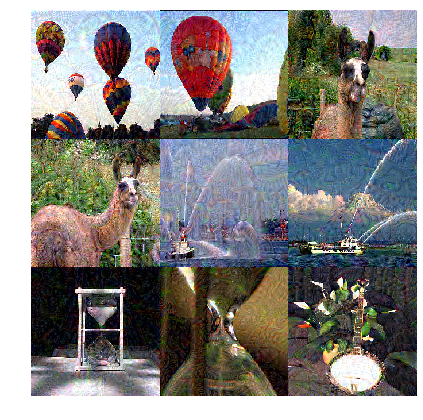} \\
 Clip range - 20, avg. transferability - 70\% & Clip range - 30, avg. transferability - 81\% \\[6pt]
 \\[6pt]
  \includegraphics[scale=0.3]{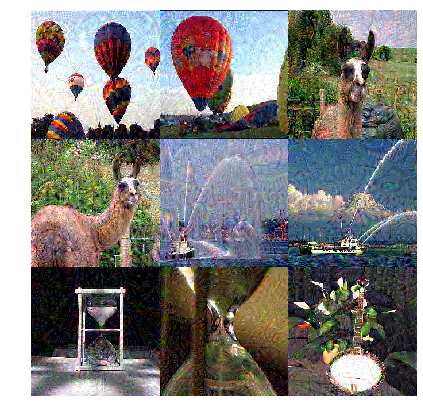} &   \includegraphics[scale=0.3]{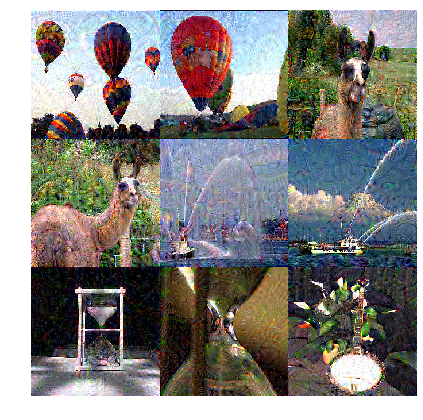} \\
 Clip range - 40, avg. transferability - 86\% & Clip range - 50, avg. transferability - 88\% \\[6pt]
 \\[6pt]
  \includegraphics[scale=0.3]{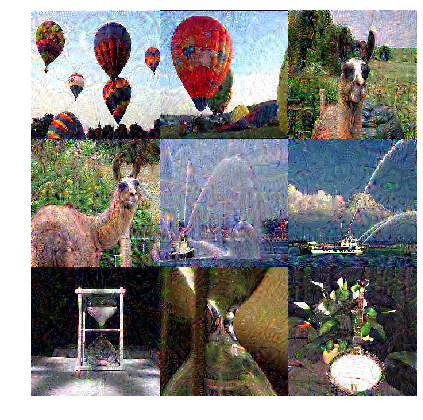} &   \includegraphics[scale=0.3]{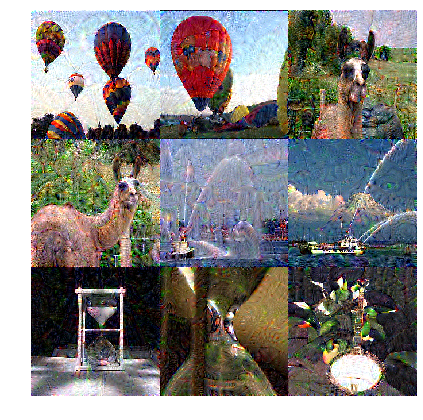} \\
Clip range - 60, avg. transferability - 89\% & Clip range - 70, avg. transferability - 89\% \\[6pt]
 \\[6pt]
\end{tabular}
\caption{Images for the I-FGSM attack trained on the VGG19 nework of different \textit{L-Infinity} clipping ranges and the average transferability(rounded) of these ranges across all models, including the source}
\label{fig: appendix2}
\end{figure}

\begin{figure}
\centering
\begin{tabular}{cc}
  \includegraphics[scale=0.3]{proj/images/orig.png} &   \includegraphics[scale=0.3]{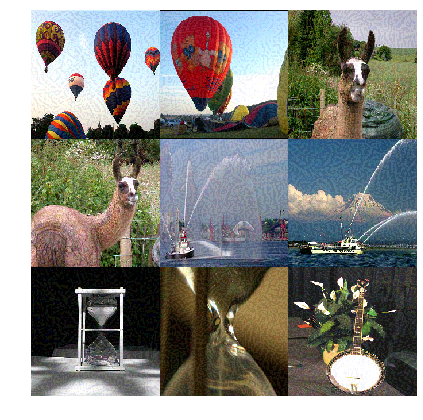} \\
 Original images & Clip range - 10, avg. transferability - 48\% \\[6pt]
 \\[6pt]
  \includegraphics[scale=0.3]{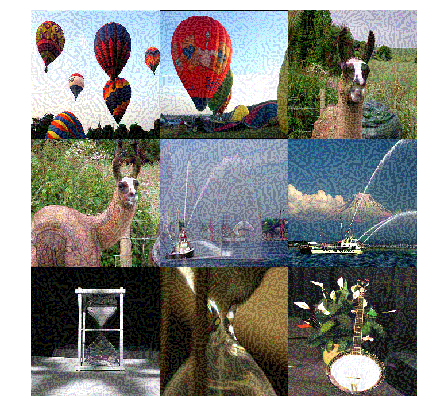} &   \includegraphics[scale=0.3]{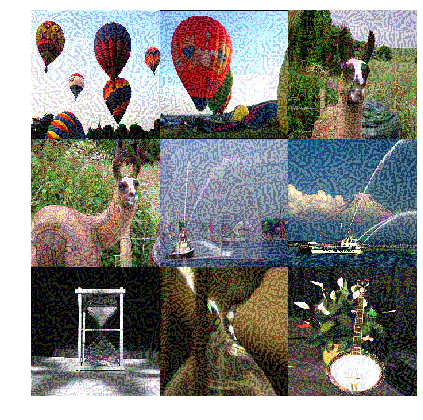} \\
 Clip range - 20, avg. transferability - 62\% & Clip range - 62\%, avg. transferability - 71\%  \\[6pt]
 \\[6pt]
  \includegraphics[scale=0.3]{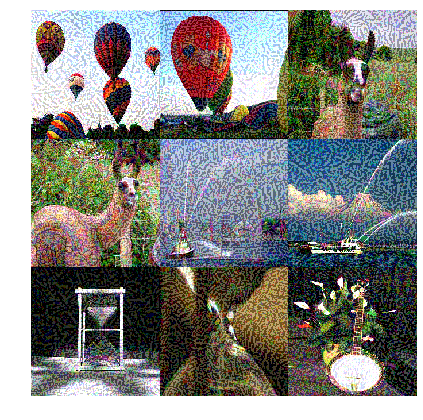} &   \includegraphics[scale=0.3]{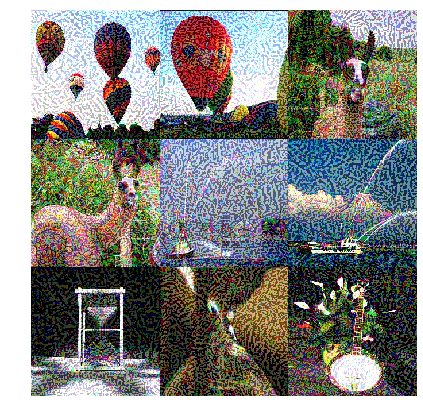} \\
 Clip range - 40, avg. transferability - 77\% & Clip range - 50, avg. transferability - 81\% \\[6pt]
 \\[6pt]
  \includegraphics[scale=0.3]{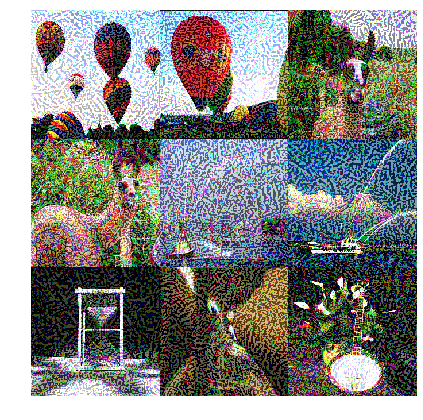} &   \includegraphics[scale=0.3]{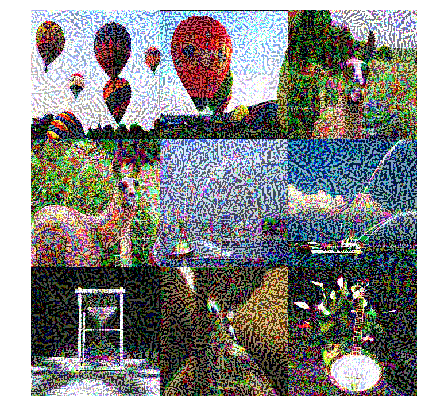} \\
Clip range - 60, avg. transferability - 85\% & Clip range - 70, avg. transferability - 88\%  \\[6pt]
 \\[6pt]
\end{tabular}
\caption{Images for the C\&W attack with LR=7, trained on the VGG19 network of different \textit{L-Infinity} clipping ranges and the average transferability(rounded) of these ranges across all models, including the source}
\label{fig: appendix3}
\end{figure}

Using \textit{SSIM} analysis(fig. \ref{fig:ssimreorg}) we are able to gather results which are more correlated with human perception than when using the \textit{L-Infinity} metric. However, due to the fact that the pixels of an image are changed in uniform ranges(figs. \ref{fig: appendix}, \ref{fig: appendix2} and \ref{fig: appendix3}), the \textit{L-Infinity} plots and the \textit{SSIM} plots are consistent with each other, although differences in performances are more pronounced in the \textit{SSIM} plots. When assessing attacks such as the Single-Pixel attack\cite{DBLP:journals/corr/abs-1710-08864} or "local" attacks, differences between the \textit{L-Infinity} and \textit{SSIM} results are more substantial. \textit{SSIM} is the superior method of evaluation of adversarial attacks to the \textit{L-Infinity} measure in the general case. However, it is still weak in evaluating certain classes of attacks such as adversarial scaling or rotations. \\

We find that, with the exception of the ensembles, for all attacks, similar architectures have similar "fooling" capacities. The adversarials found by similar architectures appear very similar as well. \\

For the FGSM attack, the most powerful and transferable model is the VGG-Ensemble, followed by VGG19/16, the Inception Ensemble, Inception Resnet V2,  and Inception V3 and Xception respectively. For the I-FGSM attack, the VGG-Ensemble is the most transferable as well for small perturbations. However, due to its quick convergence it is unable to change the image enough to to generalize to the Inception family for larger ranges of perturbation. After the VGG Ensemble, the remaining most transferable models are VGG19, VGG16, followed by the Inception family. We observe that using an ensemble of Inception models does not improve the success of the attack for both the Inception and the VGG models. For C\&W with LR=7, the ensembles perform the worst of all models. VGG19 and 16 are again the top performing models, followed by Inception Resnet V2, Xception and Inception V3. \\

\afterpage{
\begin{figure*}[p]
\centering
\subfigure[VGG16]
{\includegraphics[width=.49\columnwidth]{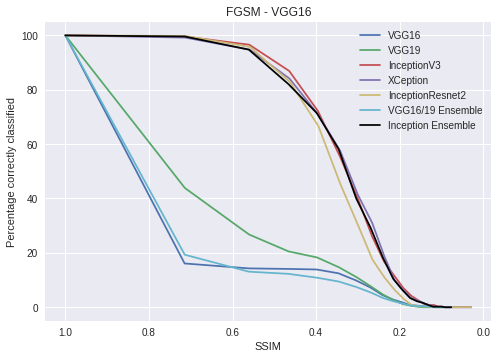}}
\subfigure[VGG19]
{\includegraphics[width=.49\columnwidth]{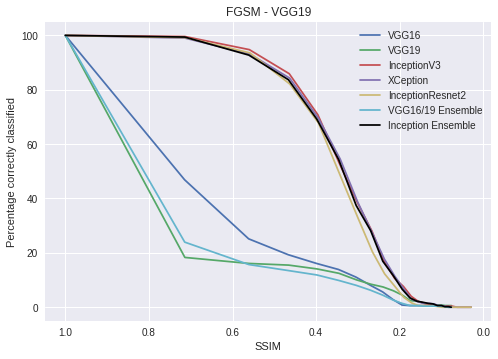}}
\subfigure[Inception V3]
{\includegraphics[width=.49\columnwidth]{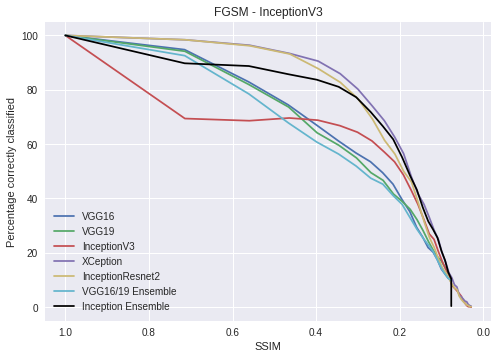}}
\subfigure[Xception]
{\includegraphics[width=.49\columnwidth]{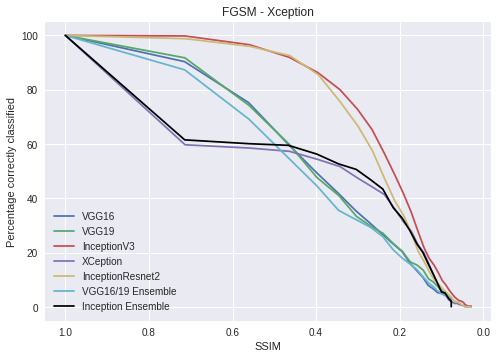}}
\vskip -0.1in
\centering
\subfigure[Inception Resnet 2]
{\includegraphics[width=.55\columnwidth]{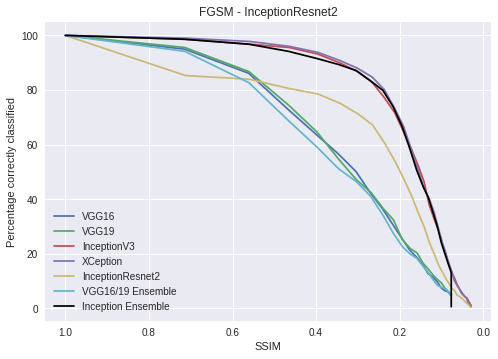}}
\subfigure[VGG Ensemble]
{\includegraphics[width=.55\columnwidth]{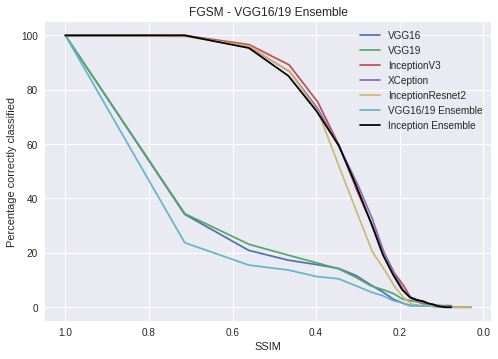}}
\subfigure[Inception Ensemble]
{\includegraphics[width=.55\columnwidth]{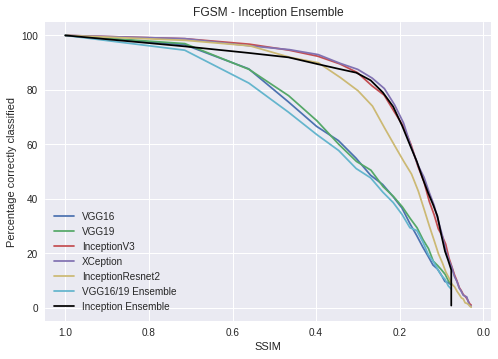}}
\subfigure[VGG16]
{\includegraphics[width=.49\columnwidth]{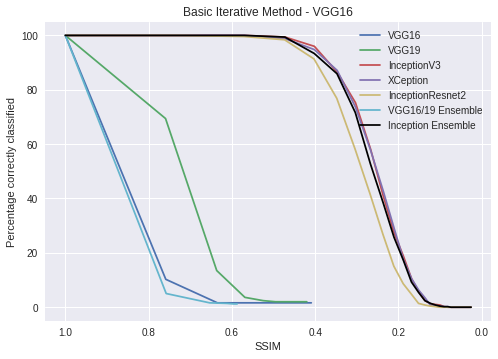}}
\subfigure[VGG19]
{\includegraphics[width=.49\columnwidth]{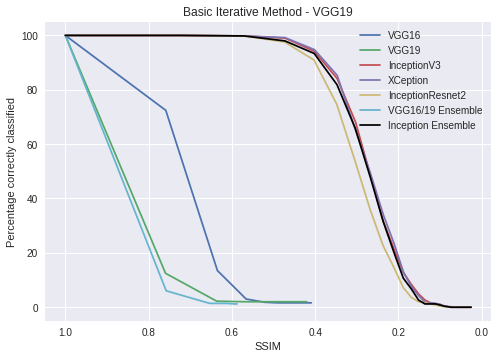}}
\subfigure[Inception V3]
{\includegraphics[width=.49\columnwidth]{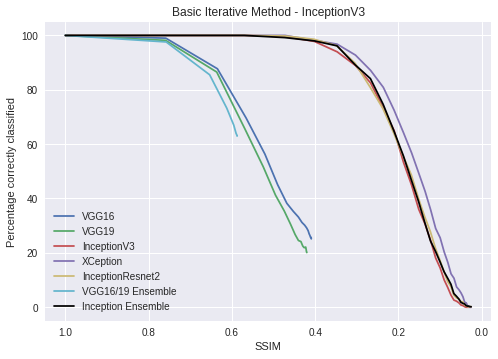}}
\subfigure[Xception]
{\includegraphics[width=.49\columnwidth]{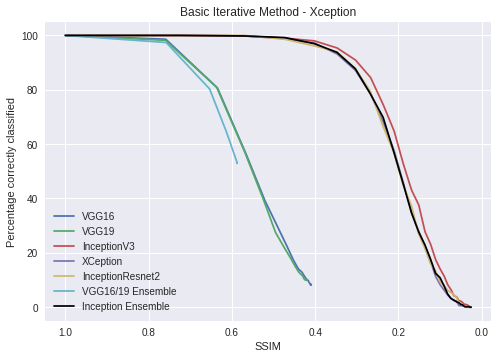}}
\vskip -0.1in
\centering
\subfigure[Inception Resnet 2]
{\includegraphics[width=.55\columnwidth]{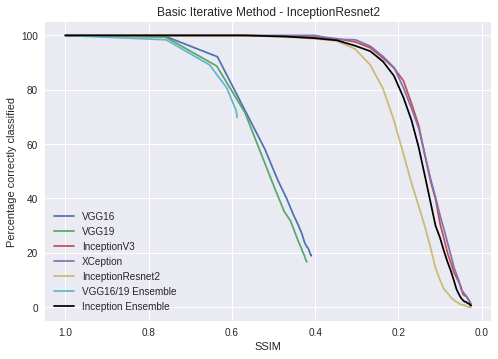}}
\subfigure[VGG Ensemble]
{\includegraphics[width=.55\columnwidth]{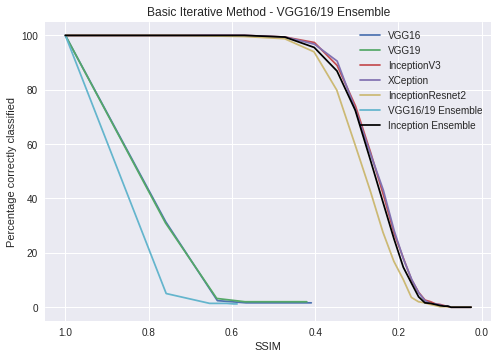}}
\subfigure[Inception Ensemble]
{\includegraphics[width=.55\columnwidth]{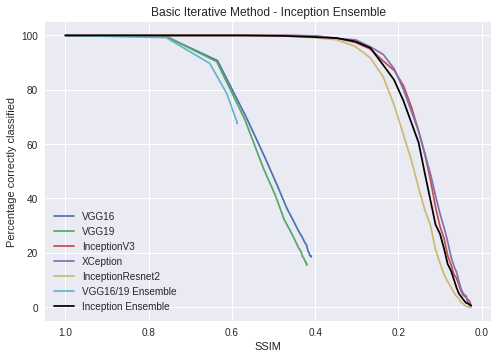}}
\subfigure[VGG16]
{\includegraphics[width=.49\columnwidth]{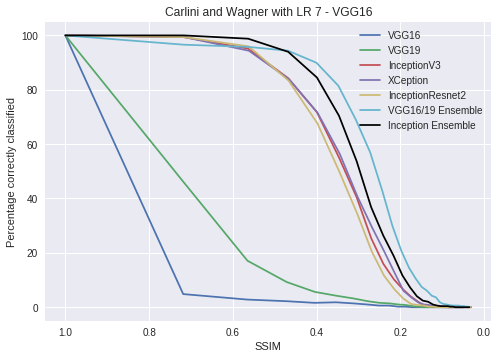}}
\subfigure[VGG19]
{\includegraphics[width=.49\columnwidth]{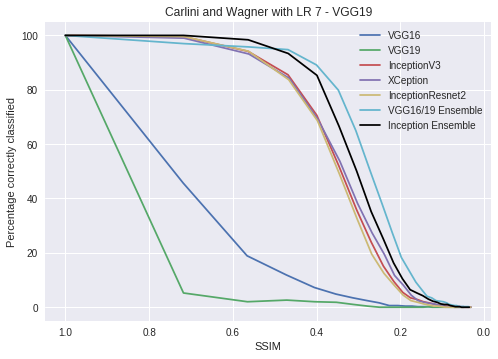}}
\subfigure[Inception V3]
{\includegraphics[width=.49\columnwidth]{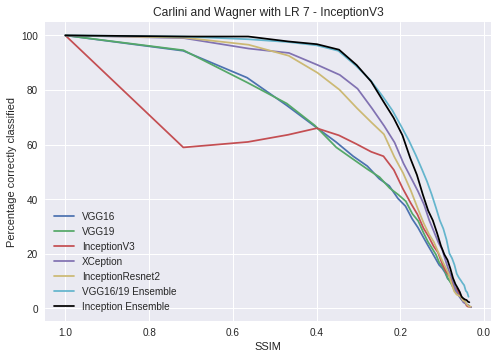}}
\subfigure[Xception]
{\includegraphics[width=.49\columnwidth]{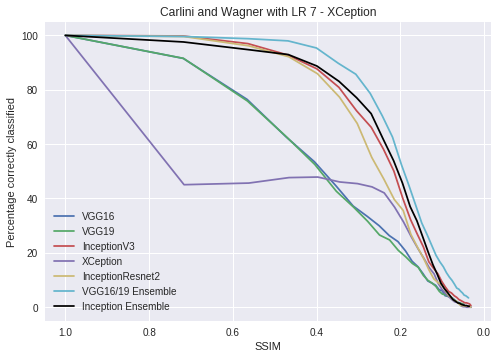}}
\vskip -0.1in
\centering
\subfigure[Inception Resnet 2]
{\includegraphics[width=.55\columnwidth]{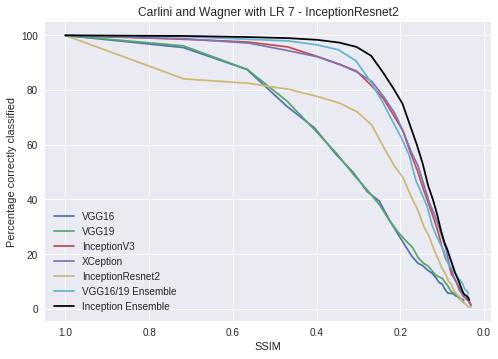}}
\subfigure[VGG Ensemble]
{\includegraphics[width=.55\columnwidth]{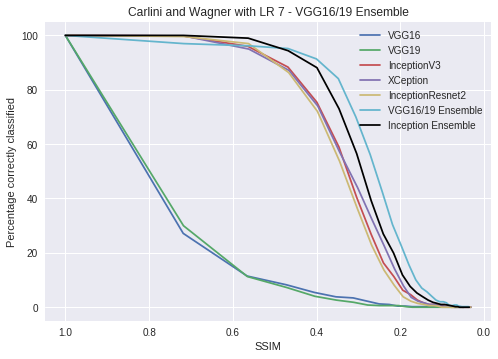}}
\subfigure[Inception Ensemble]
{\includegraphics[width=.55\columnwidth]{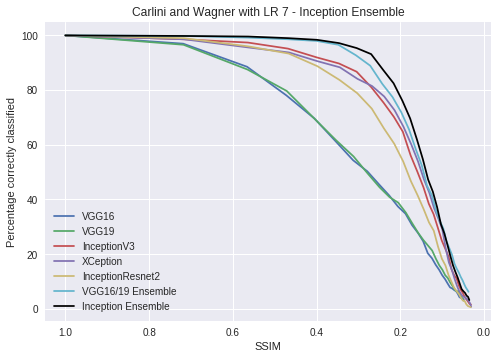}}
\vskip -0.1in
\caption{Performance of every classifier when attacked by each of the 7 classifiers, for all 3 attacks, using \textit{SSIM} to calibrate the \textit{x}-axis as metric for visual perturbation. For example, (a) shows the defensive accuracy of VGG16 when attacked by clipped adversarial images created by all 7 classifiers, using the FGSM attack(title of plot signifies the defending classifier and the used attack).}
\label{fig:ssimreorg}
\end{figure*}
\clearpage
}
In order to compare the "best-case" transferability across the attacks, we present the average transferability of the strongest attacking model for each attack across all models, including the source. \\

In order to compare the average-case transferability, results are averaged across the other models as well, and the averaged results are averaged over each attack. The \textit{SSIM} ranges of the models for each attack are averaged as well:
\begin{figure}[H]
\centering
\includegraphics[width=0.225\textwidth]{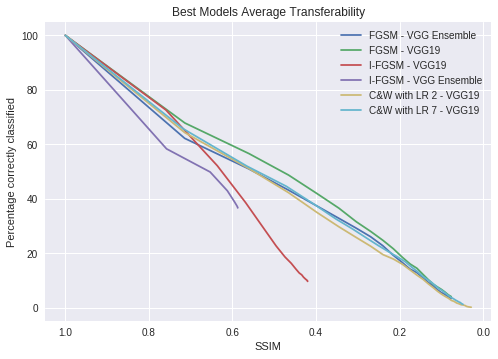}~
\includegraphics[width=0.225\textwidth]{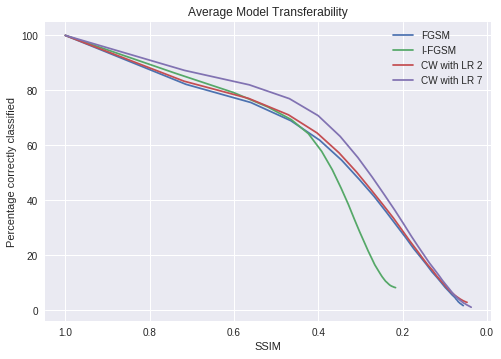}~
\caption{Comparison of the transferability of the three attacks under the chosen parameters} 
\label{fig:transferability}
\end{figure}
\section{Conclusion}
Transferable adversarial examples are a potential risk to a variety of applications employing machine learning methods by using the examples created using a local model to attack a remote service. In this work we evaluate the transferability of adversarial examples using the setting of strong untargeted attacks for the source models, which introduce high visual perturbation to the images. We measure the accuracy of the classifications of the source and target models using different sets of \textit{L-Infinity} clipped images. The results are presented with respect to both \textit{L-Infinity}, and \textit{SSIM} as metrics for visual perturbation. We find that Inception Score is not a reliable way of measuring the quality of adversarial images.We observe that for all attacks, the VGG models are a stronger attacker than the Inception models, which is consistent with findings in the literature. Moreover, ensembles have been used before in the Clarifai.com black-box attack with high success rate\cite{DBLP:journals/corr/LiuCLS16}. We find that using ensembles of parallel models with averaged predictions does not improve the attack success for all attack methods with an arbitrary parameter setting for the attack. The presented method for evaluation of transferability is not exhaustive, but significantly reduces the search space of possible parameters for the attacks by restricting them to only those which make the source misclassify most of the images, without worrying about high visual perturbation, which is amended by the clipping mechanism. Evaluating the algorithms in the presented way could lead to a more systematic and quantifiable way to compare attack tranferability than the methods currently used in the literature. More research is needed in finding ways to evaluate transferability in a consistent and fair manner.


\bibliography{references}
\bibliographystyle{icml2019}
\onecolumn
\fancyhf{}
\fancyhead[L]{\textbf{Appendix}}

\end{document}